\documentclass{article}

\PassOptionsToPackage{numbers, compress}{natbib}

\usepackage[preprint]{neurips_2025}

\usepackage[utf8]{inputenc} 
\usepackage[T1]{fontenc}    
\usepackage{hyperref}       
\usepackage{url}            
\usepackage{booktabs}       
\usepackage{amsfonts}       
\usepackage{nicefrac}       
\usepackage{microtype}      
\usepackage{xcolor}         
\usepackage{amsmath}
\usepackage{subcaption}
\usepackage{makecell}
\usepackage{multirow}
\usepackage{soul}
\usepackage{enumitem}
\usepackage{graphicx}
\usepackage{fancyvrb}
\usepackage{listings}
\usepackage{wrapfig}
\usepackage[most]{tcolorbox}
\tcbuselibrary{listingsutf8}

\newcommand{\eg}{\emph{e.g.}}

\newcommand{\etal}{\emph{et al.}}

\newcommand{\myparagraph}[1]{{\vspace{.2em} \noindent \bf #1}}

\title{Comp-X: On Defining an Interactive \\ Learned Image Compression Paradigm With Expert-driven LLM Agent}

\author{
Yixin Gao\textsuperscript{1}, Xin Li\textsuperscript{1}, Xiaohan Pan\textsuperscript{1}, Runsen Feng\textsuperscript{1}, Bingchen Li\textsuperscript{1}, \\ \textbf{Yunpeng Qi\textsuperscript{1}, Yiting Lu\textsuperscript{1}, Zhengxue Cheng\textsuperscript{2}, Zhibo Chen\textsuperscript{1}, Jörn Ostermann\textsuperscript{3}} \\
\textsuperscript{1}University of Science and Technology of China, 
\textsuperscript{2}Shanghai Jiao Tong University, \\
\textsuperscript{3}Leibniz Universität Hannover \\
\texttt{\{gaoyixin, pxh123, fengruns, lbc31415926, qiyp, luyt31415\}@mail.ustc.edu.cn}, \\
\texttt{\{xin.li, chenzhibo\}@ustc.edu.cn}, \texttt{zxcheng@sjtu.edu.cn}, \\
\texttt{ostermann@tnt.uni-hannover.de}
}

\begin{document}

\maketitle

\begin{abstract}
We present Comp-X, the first intelligently interactive image compression paradigm empowered by the impressive reasoning capability of large language model (LLM) agent. Notably, commonly used image codecs usually suffer from limited coding modes and rely on manual mode selection by engineers, making them unfriendly for unprofessional users. To overcome this, we advance the evolution of image coding paradigm by introducing three key innovations: (i) \textbf{multi-functional coding framework}, which unifies different coding modes of various objective/requirements, including human-machine perception, variable coding, and spatial bit allocation, into one framework. 
 (ii) \textbf{interactive coding agent}, where we propose an augmented in-context learning method with coding expert feedback to teach the LLM agent how to understand the coding request, mode selection, and the use of the coding tools. 
 (iii) \textbf{IIC-bench}, the first dedicated benchmark comprising diverse user requests and the corresponding annotations from coding experts, which is systematically designed for intelligently interactive image compression evaluation. Extensive experimental results demonstrate that our proposed Comp-X can understand the coding requests efficiently and achieve impressive textual interaction capability. Meanwhile, it can maintain comparable compression performance even with a single coding framework, providing a promising avenue for artificial general intelligence (AGI) in image compression.
\end{abstract}

\section{Introduction}
\label{sec:intro}
\begin{figure}[tb]
\centerline{\includegraphics[width=0.85\linewidth]{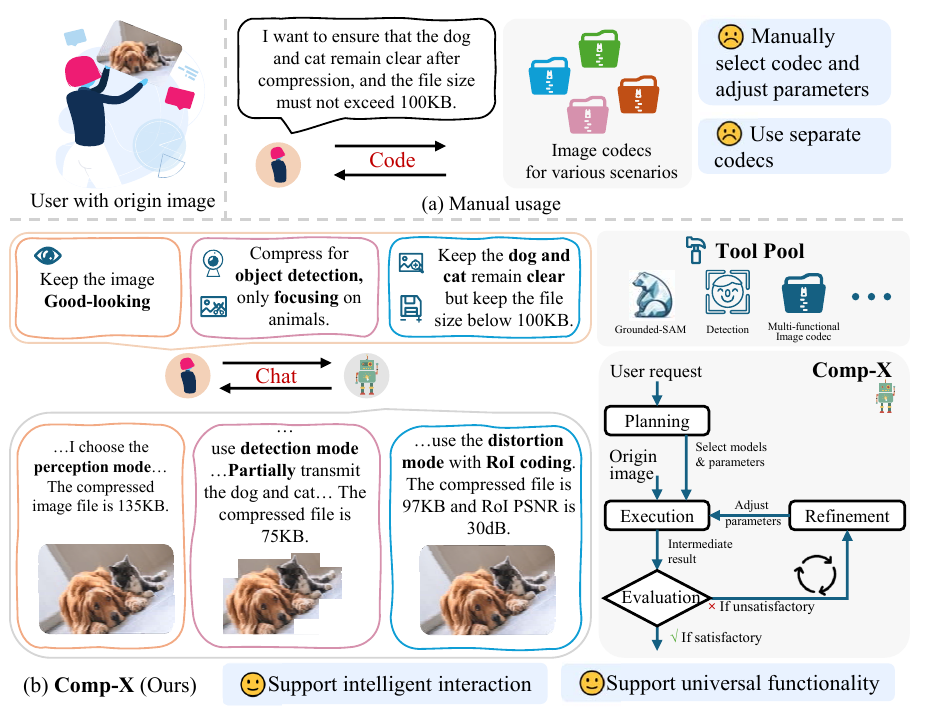}}
\caption{Image compression paradigm comparison between (a) manually inputting coding parameters and (b) our proposed Comp-X, which supports user input via natural language instructions.}
\label{fig:overview}
\vspace{-3mm}
\end{figure}
Lossy image compression is a cornerstone technology in visual signal communication, significantly reducing bandwidth and storage costs. In recent years, learned image compression (LIC) methods have achieved comparable or even superior rate-distortion performance to traditional image codecs, such as HEVC~\cite{HEVC} and VVC~\cite{VVC}, through carefully designed techniques in transform~\cite{cheng2020learned,zhu2022transformer,liu2023learned,li2023frequency,wu2021learned}, quantization~\cite{yang2020improving,guo2021soft,feng2023nvtc}, and entropy modeling~\cite{balle2018variational,minnen2018joint,minnen2020channel,guo2021causal,he2022elic}.
Besides, taking advantage of the capability of end-to-end optimization with different loss metrics, many researchers have expanded the scope of LIC to support various scenarios like region-of-interest (RoI) coding~\cite{song2021variable,kao2023transformer}, human perception~\cite{mentzer2020high,msillm,yang2024lossy,perco}, and machine perception~\cite{he2019beyond,sun2020semantic,feng2022image,feng2023semantically}. 
Despite these achievements, LIC methods typically lack a unified and user-friendly usage mechanism, limiting their accessibility for common users. Most LIC approaches remain task-specific, optimized for distinct objectives rather than offering a flexible and generalizable framework. This limitation presents two major challenges: (i) the lack of a unified framework leads to inefficiencies in real-world deployment, and (ii) common users struggle to manually select the optimal compression parameters to meet their specific requirements.

Recently, driven by the powerful reasoning capabilities of large language models (LLMs) in addressing various natural language processing tasks~\cite{gpt3,RLHF,SUPERNATURALINSTRUCTIONS, qwen,PaLM,touvron2023llama}, LLM-based agents have emerged as promising solutions to complex problems~\cite{gupta2023visual,wu2023visual,lu2024chameleon,shen2024hugginggpt,yang2024gpt4tools,llavaplus}. 
Typically, these agents employ LLMs as the core component of brain or controller and expand their perceptual and action space through strategies such as multimodal perception and tool utilization~\cite{nakano2021webgpt,yao2023react,schick2023toolformer,lu2023chameleon,wei2022chain}. For instance, HuggingGPT~\cite{shen2024hugginggpt} aims to automatically solve different user requests by prompting ChatGPT to act as a controller, conduct task planning, select suitable models in Hugging Face, and summarize responses based on execution results. 
These developments lead us to the following question: can we fundamentally evolve LIC into an interactive, user-friendly, and highly adaptable compression paradigm by leveraging the advanced reasoning and interaction capabilities of LLM agents?

In this work, we present the first intelligently interactive image compression paradigm powered by LLM agent, Comp-X, aiming to enable common users to achieve customized image compression by simply inputting natural language instructions. 
Unlike most methods where users must manually configure image codec parameters for each scenario (Fig.~\ref{fig:overview} (a)), our proposed Comp-X (Fig.~\ref{fig:overview} (b)) employs an LLM agent to interpret user instructions, automatically selecting and adjusting coding parameters and tools to fulfill diverse user requirements.

To achieve this, we first build a multi-functional image codec capable of flexibly scheduling various compression patterns, including human and machine perception, variable-rate coding, and spatial bit allocation, within a single unified framework. Specifically, we first build an image codec supporting variable bitrate (VBR) and spatial bit allocation, built upon a semantically structured bitstream~\cite{sun2020semantic,feng2023semantically}, achieving state-of-the-art rate-distortion performance. Next, to accommodate diverse user scenarios, we insert task-specific adapters~\cite{chen2022adaptformer,chen2023transtic} into our pre-trained MSE-optimized codec, fine-tuning it for perceptual quality and machine-vision tasks such as classification and segmentation. 
Based on this unified codec, we integrate an LLM agent to equip our image compression system with interactive, understanding, and reasoning capabilities. However, we observe that the pre-trained LLM~\cite{gpt4} lacks specific knowledge about image compression techniques, leading to ambiguity when interpreting user requests. To address this challenge, we propose an augmented in-context learning method with coding expert feedback (ICL-EF). 
Rather than merely providing in-context examples~\cite{incontextsurvy}, our approach involves an iterative process where a coding expert reviews the output of the agent reasoning, identifies errors, and guides the agent through corrections until accurate conclusions are achieved. 
This mechanism enables the LLM agent to effectively internalize coding-related knowledge, significantly enhancing its ability to interpret user requests accurately, select optimal coding modes, and efficiently utilize available compression tools.
Furthermore, to systematically evaluate the user request analysis capability of our Comp-X, we propose the first benchmark designed for \textbf{I}terative and \textbf{I}ntelligent image \textbf{C}ompression, called IIC-Bench. 
This benchmark contains a diverse collection of human requests, covering scenarios from simple to highly complex compression requirements. Extensive experiments demonstrate that Comp-X could accurately understand and interpret user requests, achieving an instruction parsing success rate exceeding 80\%, while maintaining competitive compression performance across diverse application scenarios.

The contributions of this paper can be summarized as follows:
\begin{itemize}[leftmargin=1em]
\item To the best of our knowledge, we present the first intelligently interactive learned image compression paradigm. By leveraging LLM agent, our approach enables an automated and customizable image coding process tailored to user inputs in natural language.
\item We build a multi-functional image codec that integrates multiple coding modes into a single unified framework, including human-machine perception, variable coding, and spatial bit allocation, eliminating the need for multiple task-specific codecs.
\item We develop an augmented in-context learning approach with coding expert feedback to enhance the understanding of image compression knowledge priors of LLM.  This mechanism significantly improves the ability of LLM agent to understand user requests, select appropriate coding modes, and effectively utilize coding tools.
\item We introduce IIC-Bench, the first dedicated benchmark designed to evaluate intelligently interactive image compression, featuring diverse real-world user requests with expert annotations, facilitating comprehensive performance assessment.
\item Extensive experiments demonstrate that Comp-X is capable of accurately interpreting user requests, achieving an instruction parsing success rate of over 80\%, while maintaining competitive compression performance across diverse application scenarios.
\end{itemize}

\section{Related Work}
\label{sec:formatting}
\subsection{Learned Lossy Image Compression}
Learned lossy image compression methods based on nonlinear transform coding \cite{balle2020nonlinear} have advanced rapidly in recent years. Early works focused on enabling end-to-end training via developing differential quantization and rate estimation techniques~\cite{balle2017end,agustsson2017soft,theis2017lossy}. Subsequently, considerable efforts have been made to design powerful neural network modules to enhance transform~\cite{cheng2020learned,zhu2022transformer,liu2023learned,li2023frequency}, quantization~\cite{yang2020improving,guo2021soft,feng2023nvtc}, and entropy model~\cite{balle2018variational,minnen2018joint,minnen2020channel,guo2021causal,he2022elic}. As a result, certain neural codecs~\cite{liu2023learned, he2022elic,guo2021causal,cheng2020learned,minnen2020channel,minnen2018joint,li2023frequency} could match or surpass the rate-distortion performance of the state-of-the-art traditional coding standards like HEVC \cite{HEVC} and VVC \cite{VVC}. 
Taking advantage of end-to-end optimization, neural image codecs have also demonstrated impressive performance in various reconstruction objectives by adjusting optimization metrics, such as high perceptual quality~\cite{high-fidelity-generative-compression,msillm,yang2024lossy,perco} and high accuracy in machine vision tasks~\cite{feng2022image,feng2023semantically,sun2020semantic,chen2023transtic}.

\subsection{Versatile Learned Image Compression}
Most neural image compression methods focus on optimizing the trade-off between compression ratio and reconstruction quality. These approaches often necessitate multiple distinct networks to satisfy diverse requirements (\eg, different bit rates), leading to substantial training costs and storage demands. 
To mitigate this problem, some previous studies have explored the versatility of neural image codecs, intending to control bitrates and reconstruction objectives within one model.
Specifically, VBR methods~\cite{toderici2015variable,choi2019variable,cui2021asymmetric,song2021variable,sun2021interpolation} attempt to achieve rate adaptation within a single model. For instance, Cui \etal~\cite{cui2021asymmetric} employ learnable channel-wise scales on latent variables for VBR implementation. Song \etal~\cite{song2021variable} propose a spatially adaptive rate adaptation scheme to enable VBR while supporting region-of-interest (ROI) coding.
Moreover, to realize distortion-perception control,
MRIC~\cite{agustsson2023multi} uses a conditional generator to trade off distortion and realism from a single representation.
While DIRAC~\cite{ghouse2023residual} and EGIC~\cite{korber2025egic} incorporate residuals into a pre-trained codec to manage the trade-off using an interpolation strategy.
In addition, CRDR~\cite{CRDR} integrates InterpCA~\cite{sun2021interpolation} into MRIC~\cite{agustsson2023multi} to achieve rate-distortion-perception control.
For distortion-cognition control, SSIC~\cite{sun2020semantic,feng2023semantically} utilizes a semantic structured bitstream to provide both high compression efficiency and functionality through selective transmission and reconstruction.
Furthermore, Liu \etal~\cite{liu2025rate} first develop a VBR codec optimized for machine vision and then establish an auxiliary branch for distortion-oriented compressed images. 

\subsection{Large language model-based agents}
AI agents are artificial entities capable of perceiving their environments and making decisions based on these perceptions to achieve specific objectives~\cite{xie2024large,xi2023rise,wooldridge1995intelligent}. 
In recent years, given the remarkable reasoning capabilities of large language models (LLMs) in addressing various natural language processing tasks~\cite{gpt3,RLHF,SUPERNATURALINSTRUCTIONS,qwen,PaLM,touvron2023llama}, many researchers have employed LLMs to conduct agents to tackle complex tasks autonomously~\cite{gupta2023visual,wu2023visual,lu2024chameleon,shen2024hugginggpt,yang2024gpt4tools,llavaplus}. In particular, LLM-based agents utilize LLMs as the central component for problem decomposition, task planning, tool usage, and human interaction. 
For example, HuggingGPT~\cite{shen2024hugginggpt} aims to automatically solve different user requests by prompting ChatGPT to act as a controller, conduct task planning, select suitable models in Hugging Face, and summarize responses based on execution results. Gpt4tools~\cite{yang2024gpt4tools} proposes a tool-related instructional dataset and then finetunes open-source LLMs to enhance tool-usage capabilities.
Nowadays, LLM-based agents have been applied to various real-world scenarios like image generation~\cite{yang2024mastering,zhao2024diffagent}, software development~\cite{li2023camel,qian2023communicative}, scientific research~\cite{boiko2023emergent,wang2024legoprover}, bringing us closer to realizing artificial general intelligence.
To the best of our knowledge, we are the first to introduce an LLM agent in learned image compression, creating an interaction-friendly image compression system.

\section{Method}
\subsection{Overview}
\begin{figure*}[tb]
\centerline{\includegraphics[width=1.0\linewidth]{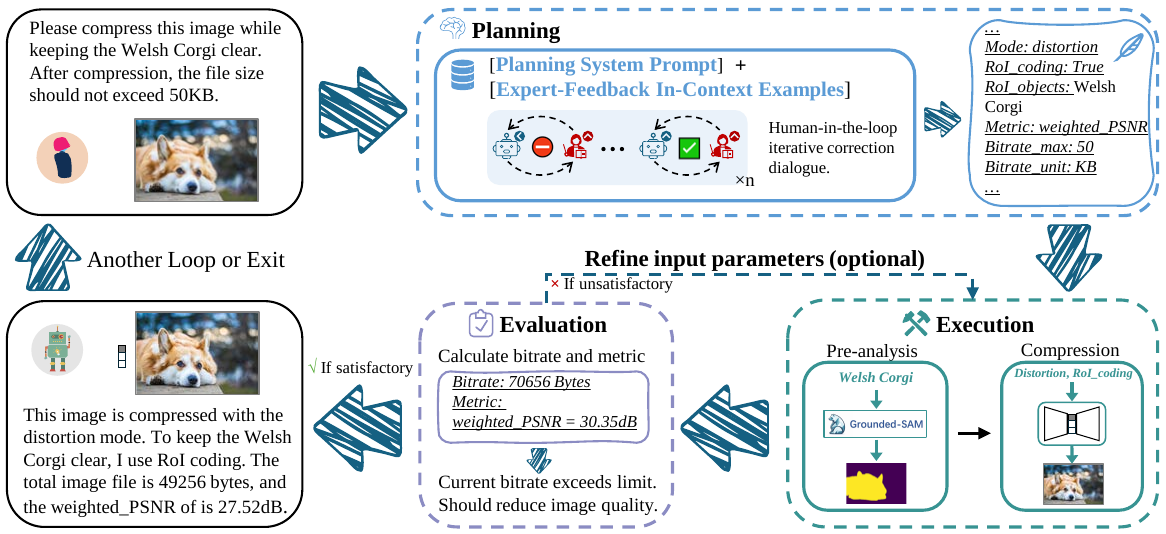}}
    \caption{Overall pipeline of our Comp-X.}
    \vspace{-3mm}
\label{fig:pipeline}    
\end{figure*}

Our Comp-X is an autonomous system designed for intelligent interaction and multi-functional image compression, integrating large language models (LLMs) with a versatile image codec. As illustrated in Fig.~\ref{fig:pipeline}, given a user request, our Comp-X processes an automatic image compression workflow, dynamically adapting to fulfill the target requirements.
The workflow includes three main stages: planning, execution, and evaluation, operating in an iterative manner. Specifically, the process begins with the planning stage, where the agent formulates a compression strategy by determining key parameters based on user requirements, such as compression mode, RoI coding, and bitrate constraints. The agent then proceeds to the execution stage, which includes pre-analysis (if necessary) followed by the compression process. After execution, the evaluation stage assesses the compressed image against predefined constraints (e.g., bitrate and weighted PSNR). If the result does not meet the requirements, the agent refines the input parameters and reattempts compression, repeating the cycle until a satisfactory outcome is achieved. Otherwise, the process concludes successfully. We introduce the functionality of our image codec in Section 3.2. For further details on each stage of our agent, please refer to Section 3.3.

\begin{figure}[tb]
    \centering
    \begin{subfigure}{0.34\textwidth}
        \centering
        \includegraphics[width=\textwidth]{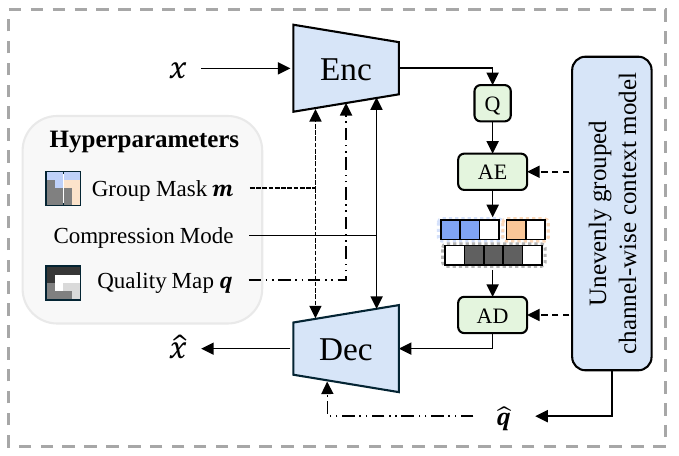}
        \caption{}
    \end{subfigure}
    \hspace{-1mm}
    \begin{subfigure}{0.45\textwidth}
        \centering
        \includegraphics[width=\textwidth]{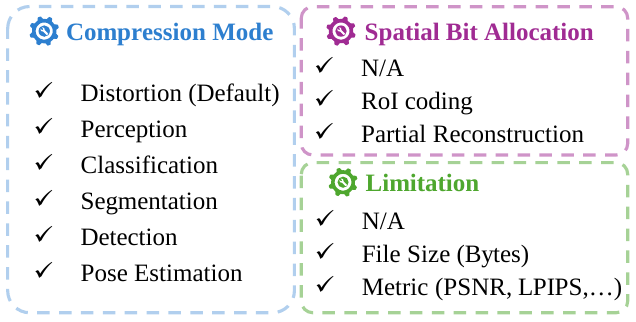}
        \caption{}
    \end{subfigure}
    \caption{Illustration of our multi-functional image codec. (a) is the high-level structure of our codec. (b) shows supportive functionality in our Comp-X.}
    \vspace{-2mm}
    \label{fig:codec_coarse}
\end{figure}

\begin{figure}[t]
    \centering
        \begin{minipage}{0.45\linewidth}
        \centering
        \includegraphics[width=\linewidth]{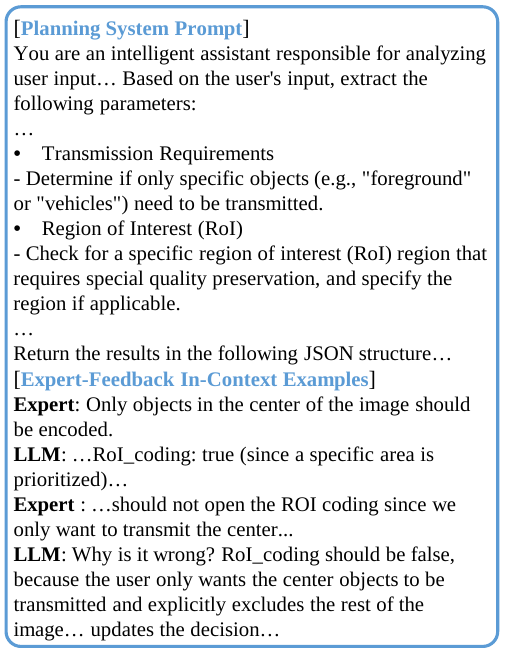}
        \caption{Prompt design for planning.}
        \label{fig:planing_prompt}
    \end{minipage}
    \begin{minipage}{0.45\linewidth}
        \centering
        \includegraphics[width=\linewidth]{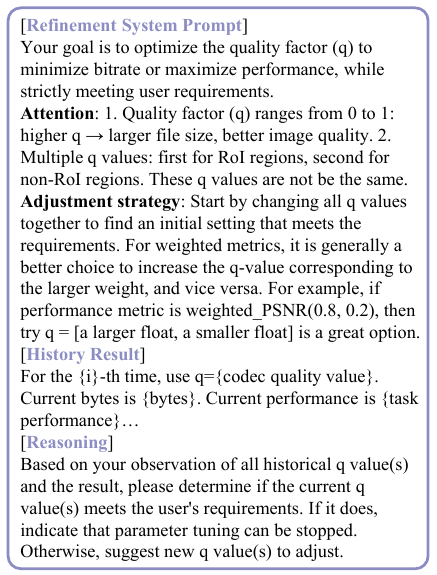}
        \caption{Prompt design for refinement.}
        \label{fig:prompt_for_refinement}  
    \end{minipage}
    \vspace{-5mm}
\end{figure}
\vspace{-5mm}

\subsection{Multi-functional Image Codec}
Modern applications increasingly demand image codecs that are flexible, efficient, and adaptive to diverse scenarios, such as human-machine interaction and machine vision. Rather than employing separate models for distinct tasks, we propose a multi-functional image codec that integrates various functionalities within a unified framework. 
Specifically, our codec builds upon and generalizes semantically structured image compression~\cite{sun2020semantic,feng2023semantically}, enabling universal and efficient usage across multiple scenarios. The key lies in decoupling the bitstream into non-overlapping object segments, facilitating intelligent tasks and human-machine interaction through selective partial transmission and reconstruction. As illustrated in Fig.~\ref{fig:codec_coarse} (a), the proposed codec employs three conditional inputs: a group mask $m$, a quality map $q$, and a task identifier (task id). 
The group mask~\cite{feng2023semantically} guides the partitioning of the latent representation into distinct object-level segments following a one-pass downsampling transform. This allows the bitstream to be efficiently organized at the object level, supporting versatile downstream applications from a single encoded stream through flexible partial transmission and reconstruction. 
The quality map, matching the spatial dimensions of the input image, modulates the bitrate at each spatial location with values ranging between 0 (lowest bitrate) and 1 (highest bitrate). Finally, the task id explicitly specifies the desired application scenario, such as distortion, perception, and segmentation. This is achieved by fine-tuning adapters~\cite{chen2022adaptformer,chen2023transtic} based on our MSE-optimized base codec. By jointly adjusting three hyperparameters, our unified codec covers three common categories of user requirements: application scenario, spatial bit allocation, and limitations, as depicted in Fig~\ref{fig:codec_coarse} (b). Further implementation details are provided in the appendix.

\subsection{Interactive Image Compression Paradigm}
\myparagraph{In-context learning with expert feedback (ICL-EF)}\quad
To help the agent understand which codec configuration matches specific application scenarios, we first define its task through a \textit{planning system prompt}, which clarifies the meaning of relevant coding parameters, their default settings, and the expected output format. To further improve the coding-related knowledge understanding of agent, we propose an in-context learning method with expert feedback (ICL-EF). In-context learning (ICL) is a widely used strategy enabling LLMs to perform a target task without gradient update~\cite{gpt3,meade-etal-2023-using,zhu2023multilingual,khorashadizadeh2023exploring,ding-etal-2023-gpt,min-etal-2022-metaicl,genartist,yang2024mastering}, which requires a prompt context containing a few demonstration examples written in natural language templates. Based on this, we introduce an additional step involving an expert feedback mechanism. 
Specifically, we invite a coding expert to evaluate the performance of understanding various user requests of LLM through multi-turn dialogues. Whenever reasoning errors are identified, the expert explicitly highlights them and guides the LLM to reconsider and revise its decisions accordingly.  Finally, the entire dialogue, including corrections and explanations, is recorded and integrated into the prompt as expert-feedback in-context examples.

An illustrative example is shown in Fig.~\ref{fig:planing_prompt}, where the expert aims to evaluate whether the LLM can accurately determine whether the user only intends to transmit a specific part of the image. Initially, the LLM agent incorrectly suggested activating RoI coding, which was unnecessary for this scenario. Following expert intervention and corrective feedback, the agent reconsidered its reasoning and revised its decision appropriately. We select several representative examples that collectively cover all major application scenarios to serve as context. In this way, we effectively supplement the pre-trained LLM with domain-specific compression knowledge while incurring negligible annotation costs.

\myparagraph{Execution mechanism}\quad
After analyzing the user request, the agent initiates the execution stage, performing the necessary actions according to the identified coding mode and specified parameters. This stage involves two main steps: pre-processing (when required) and compression. Specifically, if the chosen coding mode involves region-of-interest (RoI) coding or partial image transmission, our Comp-X automatically activates appropriate pre-processing tools, such as Grounding-SAM~\cite{ren2024grounded} and detectron2~\cite{wu2019detectron2}, to generate spatial segmentation masks. 
These masks directly correspond to quality maps and group masks required in subsequent encoding steps. 
If region-based coding is not required, the agent bypasses pre-processing and proceeds directly to the compression step. During the compression step, the agent inputs the original image, quality map, task identifier, and group masks into our multi-functional image codec to produce the compressed bitstream.

\myparagraph{Evaluation and refinement}\quad 
Once the user has customized, fine-grained requirements regarding coding efficiency, such as specific constraints on the encoded file size or decoded image quality, the coding system should be able to monitor and assess the discrepancy between the current coding outcome and user-defined limit, adjusting the codec input parameters adaptively. 
To achieve this, we design an adaptive self-refinement scheme through prompt engineering~\cite{shen2024hugginggpt,yang2024octopus,mu2023embodiedgpt}. As illustrated in Fig.~\ref{fig:prompt_for_refinement}, we first provide a clear role definition and task instructions through a carefully constructed \textit{system prompt}. 
This prompt explicitly outlines the responsibilities of the LLM agent, clarifies the practical significance of the quality factor within the image codec, and details specific parameter adjustment strategies. 
Subsequently, historical coding outcomes (\textit{history result}) are supplied, enabling the agent to analyze and exploit the intrinsic relationship between the quality factor and its corresponding bitrate or task-specific performance metric. Finally, the agent assesses whether current codec parameters satisfy user requirements. If not, it will suggest parameter adjustments to achieve optimal performance that aligns with the targeted coding goals through reasoning.


\section{Experimental Results}

\subsection{Experiment Setup}
\myparagraph{Implementation details}\quad
In our experiments, we explore employing GPT-4o~\cite{gpt4} and DeepSeek-v3~\cite{deepseekai2024deepseekv3technicalreport} as the main models in our agent, as they are two of the most representative closed-source and open-source LLMs, respectively. We train our codec on Imagenet v6 trainset~\cite{kuznetsova2020open} and COCO 2017 training set~\cite{coco}. Each batch contains 8 random $256\times256$ crops from the training dataset. 
We set the temperature of LLM to 0.7 (the default setting in GPT). All experiments with this temperature are repeated three times and the average is reported. For experiments on refinement, we set the maximum update time of our Comp-X to 10 to avoid excessively long encoding times. Further details are provided in the appendix.

\myparagraph{Baseline methods}\quad
To evaluate the effectiveness of our Comp-X, we selected a diverse set of baseline methods for human vision comparisons, including both traditional~\cite{VVC,bellard_bpg} and learned image coding methods~\cite{he2022elic,cheng2020learned,minnen2018joint,feng2023semantically,high-fidelity-generative-compression,zou2022devil}. Among these, HiFiC~\cite{high-fidelity-generative-compression} is optimized for high-fidelity reconstruction. For machine vision comparisons, we also include recent task-specific methods such as the prompt-based TransTIC~\cite{chen2023transtic} and the adapter-based Adapt-ICMH~\cite{adapt-icmh} as baselines.

\myparagraph{Evaluation datasets and protocol}\quad
For human vision tasks, we use the widely used Kodak dataset~\cite{kodak} for distortion evaluation and DIV2K validation set\cite{div2k} for perception evaluation. 
For machine vision tasks, we conduct experiments on object detection, instance segmentation and pose estimation using MS-COCO 2017 validation set~\cite{coco}, while classification tasks are evaluated on the ImageNet dataset~\cite{deng2009imagenet}. For region-of-interest (RoI) reconstruction, the image reconstruction quality is measured in terms of weighted PSNR, where the weighted mean-squared error (MSE) is:
\vspace{-0.5mm}
\begin{equation}
\text{weighted MSE} = \frac{\alpha \cdot \text{MSE}_{\text{RoI}} + \beta \cdot \text{MSE}_{\text{non-RoI}}}{\alpha \cdot N_{\text{RoI}} + \beta \cdot N_{\text{non-RoI}}},
\end{equation}
where $\alpha$ and $\beta$ are the weighting factors for ROI and non-RoI regions. $\text{MSE}_{\text{RoI}}$ and $\text{MSE}_{\text{non-RoI}}$ represent the mean squared errors within RoI and non-RoI regions. $N_{\text{RoI}}$ and $N_{\text{non-RoI}}$ denote the number of pixels in RoI and non-RoI regions. We conduct evaluations on 13 images from the Kodak dataset with distinct foreground objects, using IS-Net~\cite{qin2022} to extract segmentation masks.

\subsection{User Request Analysis}\quad

\myparagraph{IIC-Bench} We present a benchmark for \textbf{I}nteractive and \textbf{I}ntelligent image \textbf{C}ompression, termed IIC-Bench. This benchmark contains 195 items, covering simple to complex user requirements for image compression. We first collect requests from some experts and general users to ensure the diversity of dataset, then manually annotate some of them. Next, since labeling each data is heavy, we employ GPT-4o for automated labeling via in-context learning, followed by human checks and corrections. Tasks within the benchmark are categorized into simple (93 items) and hard (102 items). Simple task refers to cases where the user has only a single objective, such as specifying a application scenario or reducing file size. In contrast, hard task involves multiple user requirements simultaneously, such as prioritizing the quality of specific objects while ensuring the file size remains below 100KB.
More details about this benchmark are in the appendix.
Using this benchmark, we evaluate the success rate of user requirement analysis performed by our Comp-X.

\myparagraph{Success rate}\quad
\begin{table}[b]
    \centering
    \begin{minipage}{0.47\textwidth}
    \resizebox{\textwidth}{!}{
    \begin{tabular}{lccc}
        \toprule
        Method & Simple (\%) & Hard (\%) & All (\%) \\
        \midrule
        GPT-4o + system prompt & 70.67 & 60.46 & 65.30 \\
        \midrule
        + ICL & 74.67 & 78.10 & 77.26 \\
        + ICL-EF (Ours) & \textbf{83.87} & \textbf{81.70} & \textbf{82.74} \\
        \bottomrule
    \end{tabular}}
    \end{minipage}
    \hfill
    \begin{minipage}{0.51\textwidth}
    \resizebox{\textwidth}{!}{
    \begin{tabular}{lccc}
        \toprule
        Method & Simple (\%) & Hard (\%) & All (\%) \\
        \midrule
        DeepSeek-v3 + system prompt & 72.33 & 62.67 & 67.28 \\
        \midrule
        + ICL & 82.00 & 84.00 & 83.04 \\
        + ICL-EF (Ours) & \textbf{85.33} & \textbf{84.67} & \textbf{84.99} \\
        \bottomrule
    \end{tabular}}
    \end{minipage}
    \vspace{1mm} 
    \caption{User request analysis success rate on our IIC-bench. ICL denotes in-context learning, and ICL-EF represents our proposed in-context learning with expert feedback.}
    \label{tab:ablation_ICL_EF}
\end{table}
Table~\ref{tab:ablation_ICL_EF} presents the success rates of various methods on our IIC-Bench. Using only LLM with a system prompt yields limited results on both simple and hard tasks. We observe that this is due to the insufficient understanding of image coding priors, such as the purpose of different compression modes and the relationship between size level and quality. For example, if a user desires the highest possible visual quality, the encoding bitrate should be maximized accordingly. 

Incorporating in-context learning (ICL) significantly enhances performance, especially for analyzing complex user requests. Since challenging tasks are often requested by coding experts familiar with codec functionalities, their precise formulations can help reduce the difficulty of decision-making. However, in cases involving vague or general user requests (e.g., "as similar as possible to the original image"), ICL does not always aid the LLM in making accurate judgments. Our proposed approach, ICL-EF (in-context learning with expert feedback), consistently achieves success rates above 80\%, demonstrating significantly improved adaptability to diverse user demands. By incorporating historical dialogues containing human expert corrections, the LLM can reference accurate, compression-related reasoning logic. These findings underscore the effectiveness of integrating expert feedback into in-context learning for accurately understanding and addressing intricate user requirements.

\subsection{Main Results}
\begin{table*}[tbp]
    \centering
    \small
    \resizebox{1.0\textwidth}{!}{
    \begin{tabular}{lp{11cm}}
    \toprule
         Task &  Input User Instruction \\
         \midrule
       Single application scenario  & \textit{Compress xx.jpg for \{distortion, perception, classification, pose estimation, detection, segmentation\} task. Target a \{minimum, small, medium, large, maximum\} file size.} \\
      Multiple application scenario & \textit{Compress xx.jpg for both human vision and \{pose estimation, segmentation\} task. Target a \{minimum, small, medium, large, maximum\} file size.} \\
       Refinement: limited rate & \textit{Compress xx.png. Target a size of about xx Bytes.} \\
       Refinement: limited Performance & \textit{Compress xx.png. Target a PSNR of about xx dB.} \\
       Refinement: limited rate with RoI coding & \textit{Compress xx.jpg. Keep foreground objects with high quality. When evaluating the result, I want to use weighted PSNR and set the RoI region scale to 0.8. The target file size is about xx Bytes.} \\
       \bottomrule
    \end{tabular}}
    \caption{User instructions for image compression in our Comp-X.}
    \label{tab:test_instruction}
\end{table*}
\begin{figure*}[t]
    \centering
    \begin{minipage}{0.333\textwidth}
        \centering
        \includegraphics[width=\linewidth]{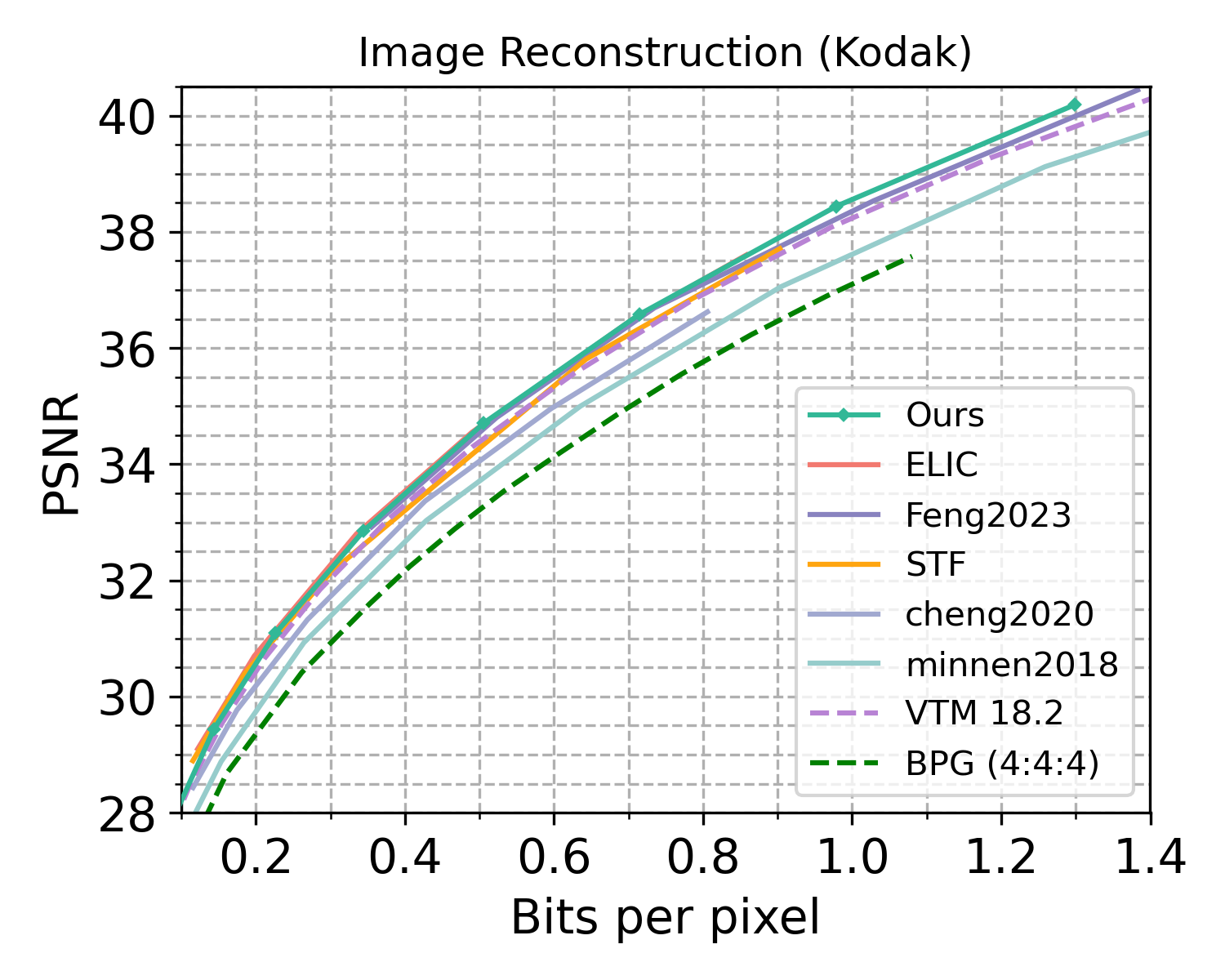}
        \hspace{-4mm}
    \end{minipage}%
    \begin{minipage}{0.333\textwidth}
        \centering
        \includegraphics[width=\linewidth]{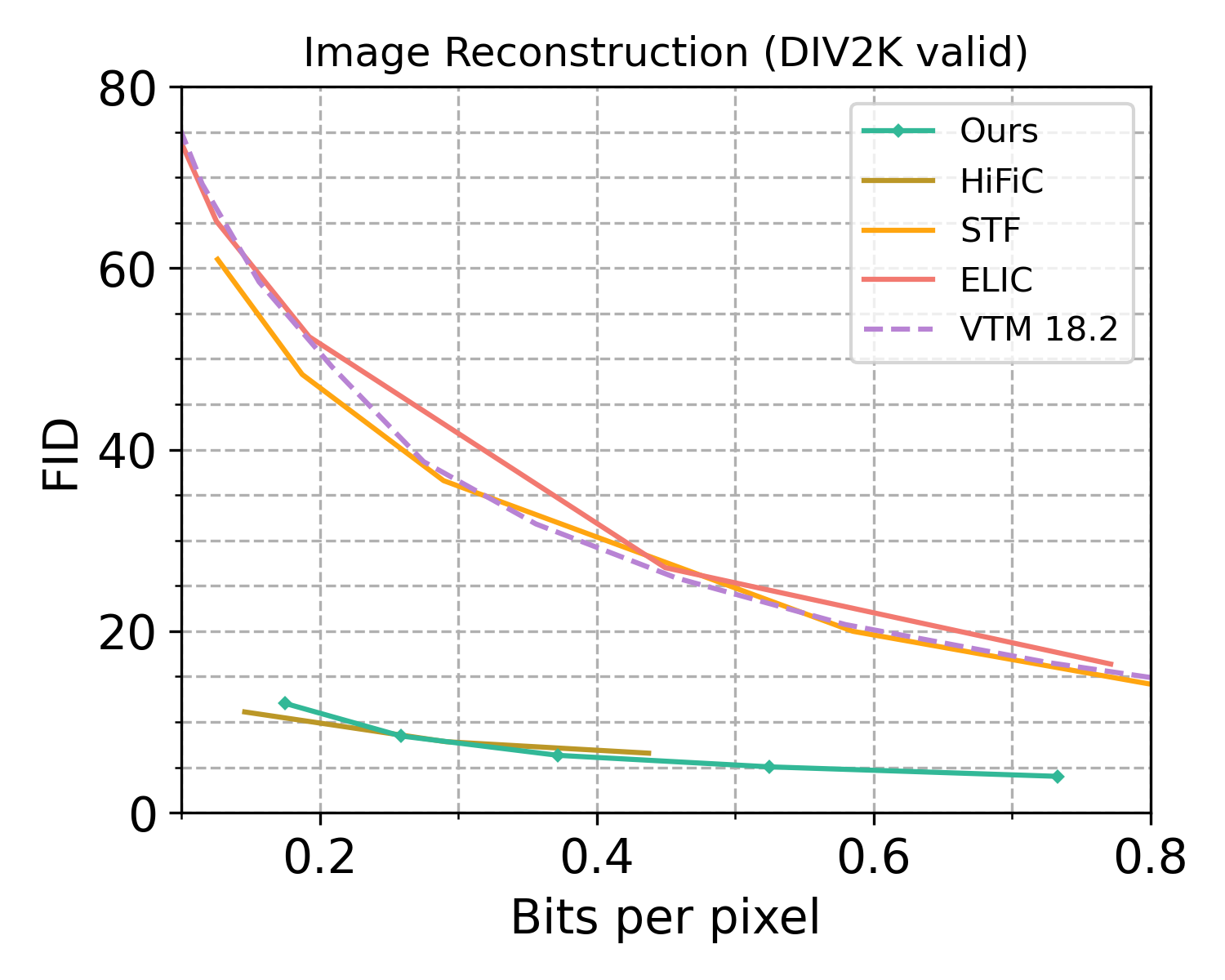}
    \end{minipage}%
    \begin{minipage}{0.333\textwidth}
        \centering
        \hspace{-4mm}
        \includegraphics[width=\linewidth]{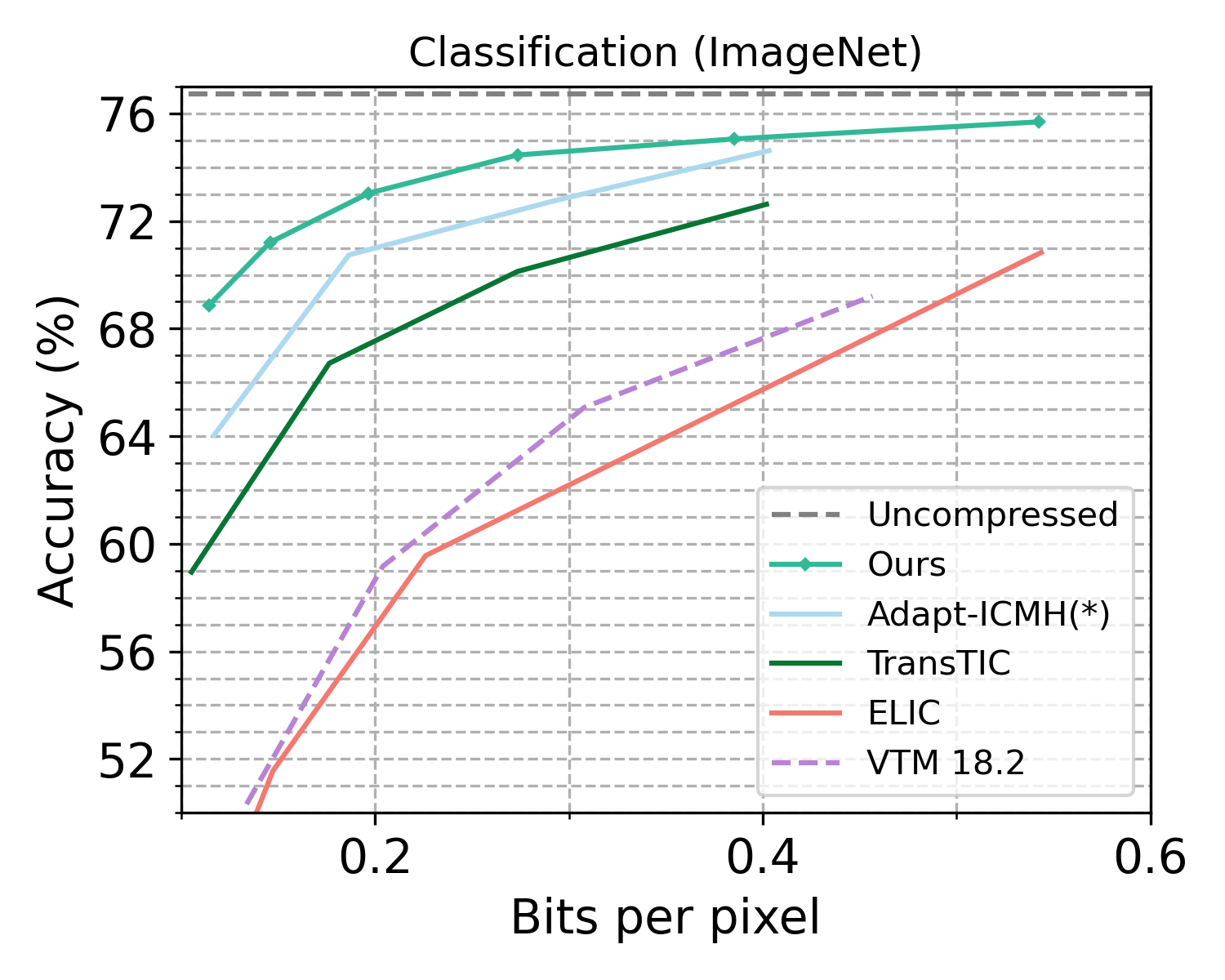}
    \end{minipage}\\[-5pt]
    \begin{minipage}{0.333\textwidth}
        \centering
        \includegraphics[width=\linewidth]{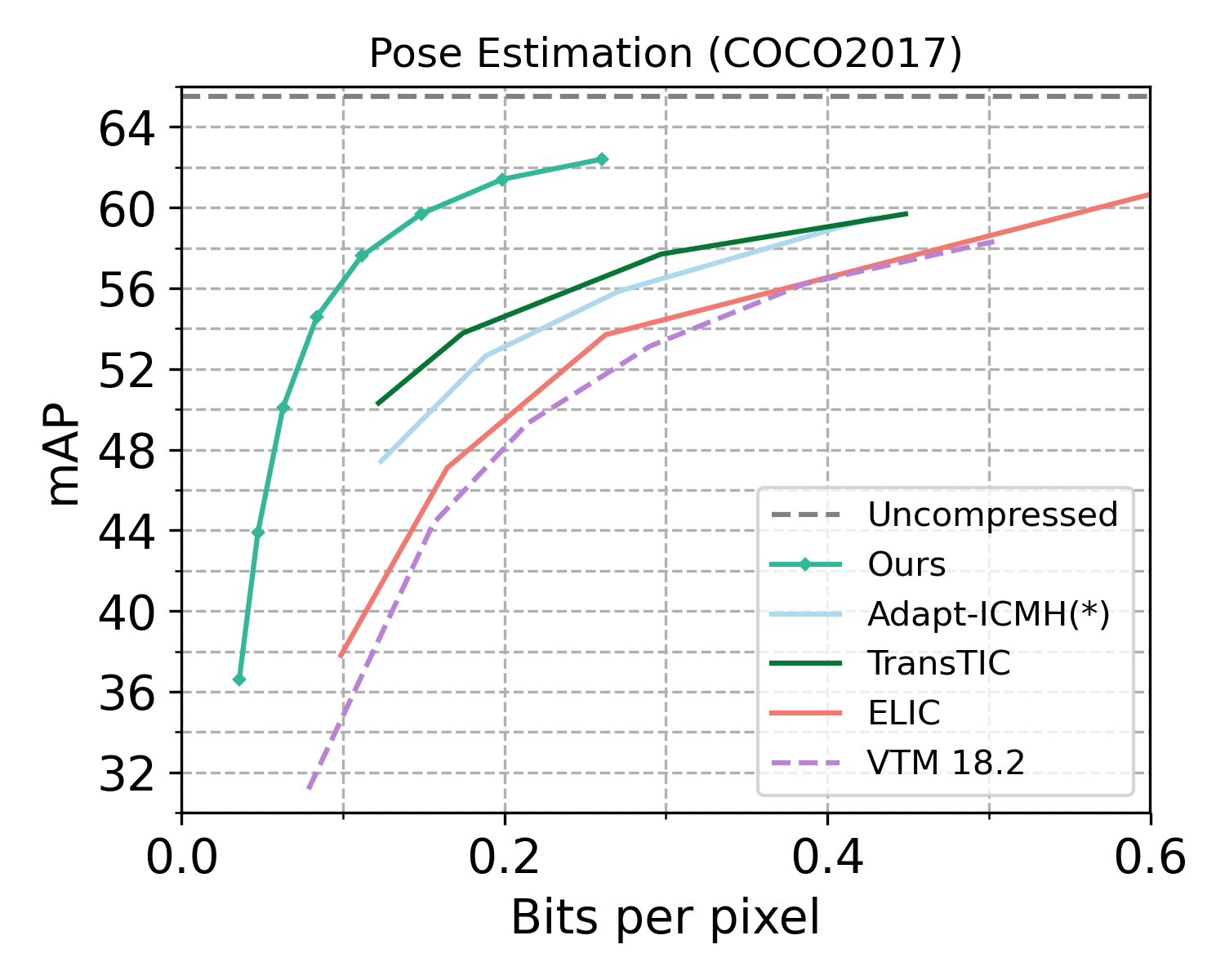}
        \hspace{-4mm}
    \end{minipage}%
    \begin{minipage}{0.333\textwidth}
        \centering
        \includegraphics[width=\linewidth]{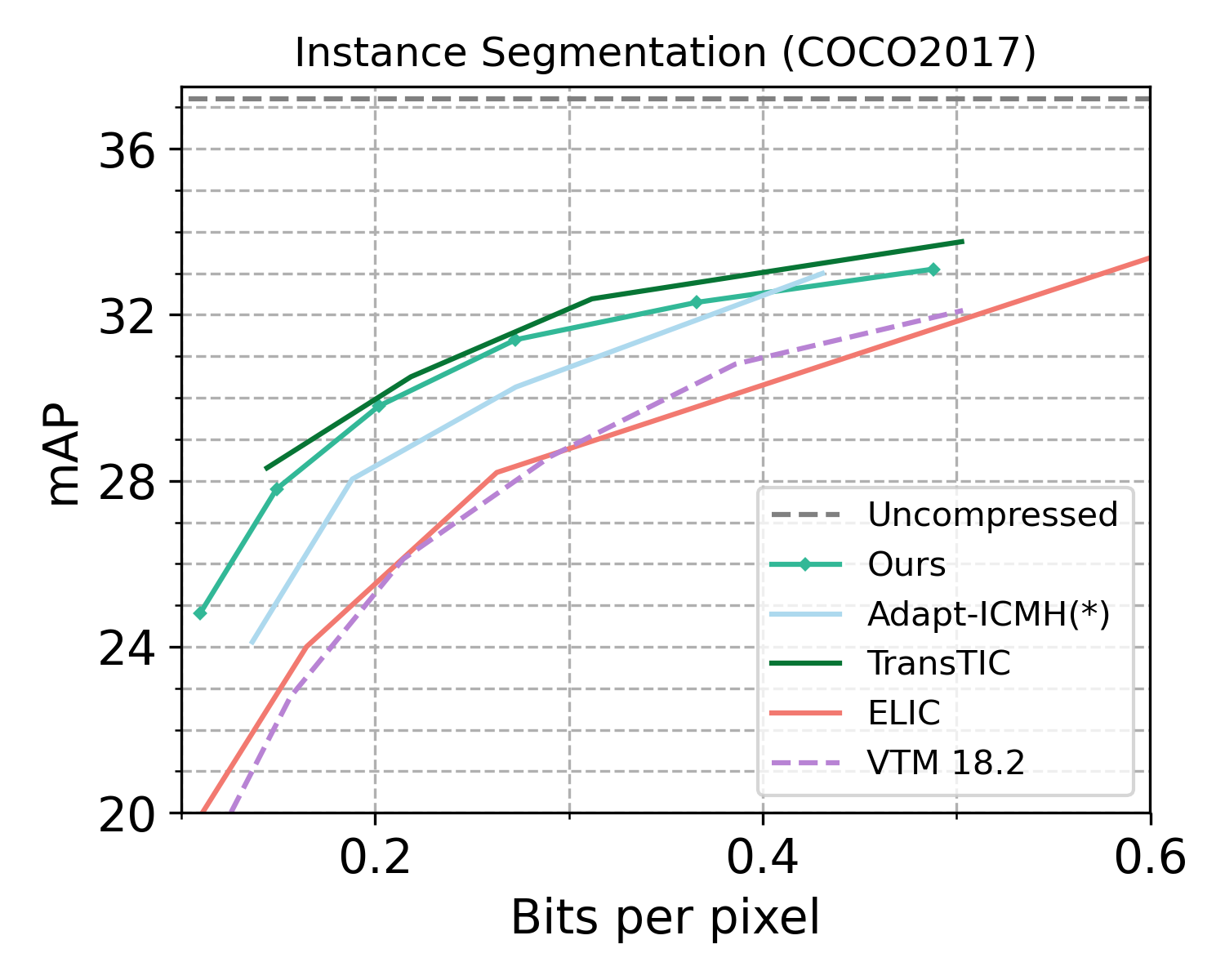}
    \end{minipage}%
    \begin{minipage}{0.333\textwidth}
        \centering
        \hspace{-4mm}
        \includegraphics[width=\linewidth]{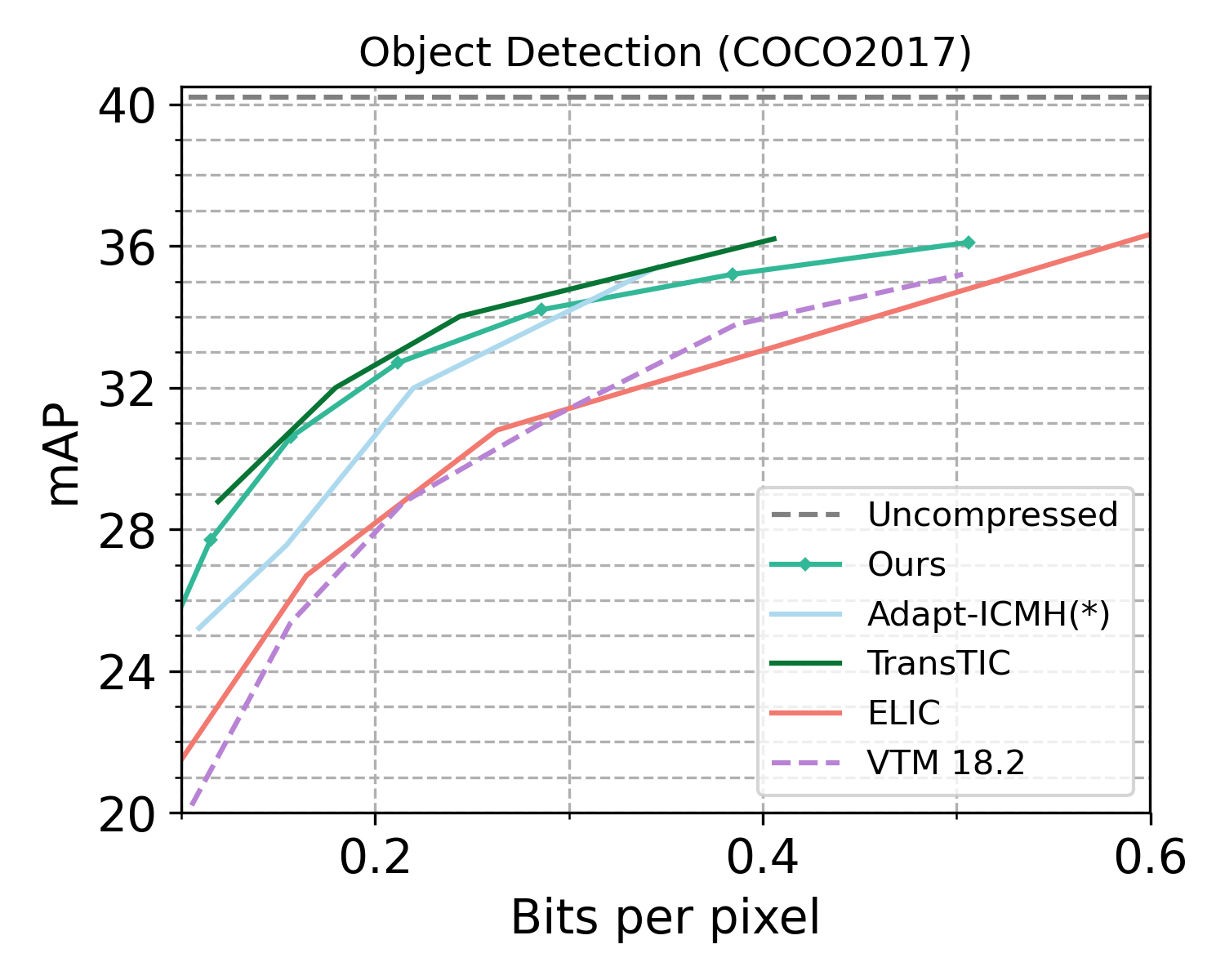}
    \end{minipage}
    \vspace{-1mm}
    \caption{Comparisons on multiple tasks. *  means our implementation.}
    \label{fig:result_tasks}
\end{figure*}
\myparagraph{Single application scenario}\quad
Fig.~\ref{fig:result_tasks} shows the rate-distortion performance of our Comp-X on six representative tasks, including distortion, perception and four machine vision tasks. We evaluate the system using user instructions provided in the first line of Table~\ref{tab:test_instruction}. Given the instruction, our Comp-X can automatically analyze and invoke the corresponding compression mode as well as apply compression at the specified bitrate levels. 
As shown in Fig.~\ref{fig:result_tasks}, considering compression efficiency for human vision, our Comp-X achieves competitive performance with state-of-the-art (SOTA) codecs in terms of PSNR and FID. In addition, Comp-X shows strong results in machine vision tasks, achieving superior performance in classification and pose estimation. For instance segmentation and object detection, our method outperforms most baseline approaches. This is due not only to task-specific fine-tuning but also to the significant bitstream savings made by semantic bitstream structuring. Through testing in six different scenarios, the results demonstrate the robustness of our Comp-X in handling single-task user cases.

\myparagraph{Multiple application scenarios}\quad 
In real-world applications, users may want to verify machine-generated decisions through reconstructed images, necessitating a balance between human and machine vision requirements. Recently, semantic structured bitstream~\cite{sun2020semantic} have enabled flexible object transmission by decomposing the bitstream at the semantic level into spatially non-overlapping segments, allowing task-specific selection of relevant objects. Our method simultaneously supports human visual perception and intelligent task analysis by leveraging this property. 
\begin{wraptable}{r}{0.5\linewidth}
    \centering
    \resizebox{0.48\textwidth}{!}{
    \begin{tabular}{lcccc}
        \toprule
        \multirow{2}{*}{Method} & \multicolumn{2}{c}{Pose Estimation} & \multicolumn{2}{c}{Segmentation} \\  
        \cmidrule(lr){2-3} \cmidrule(lr){4-5}
         & BD-PSNR $\uparrow$ & BD-mAP $\uparrow$ & BD-PSNR $\uparrow$ & BD-mAP $\uparrow$ \\
        \midrule
        VTM 18.2 & 0.00 & 0.00 & 0.00 & 0.00\\
        \midrule
        ELIC & -0.04 & 1.77 & 0.17 & 0.57 \\
        TransTIC & -2.60 & 5.85 & -2.07 & \textbf{3.72} \\
        Ours & \textbf{4.56} & \textbf{12.55} & \textbf{2.15} & 1.83 \\
        \bottomrule
    \end{tabular}}
    \caption{Comparison on BD-PSNR and BD-mAP of different codecs relative to VTM 18.2.}
    \label{tab:bd_psnr_vs_bd_map}
\end{wraptable}
As illustrated in Table.~\ref{tab:bd_psnr_vs_bd_map}, our Comp-X effectively balances the requirements of both human and machine vision. We observe that compared to ELIC, our method achieves better bitrate savings under both human and machine vision metrics when evaluating pose estimation and segmentation tasks. Furthermore, compared with task-specific optimized TransTIC, our method could ensure higher reconstruction quality for human application.

\begin{figure}[h]
    \centering
    \begin{minipage}{0.35\linewidth}
        \centering
        \includegraphics[width=\linewidth]{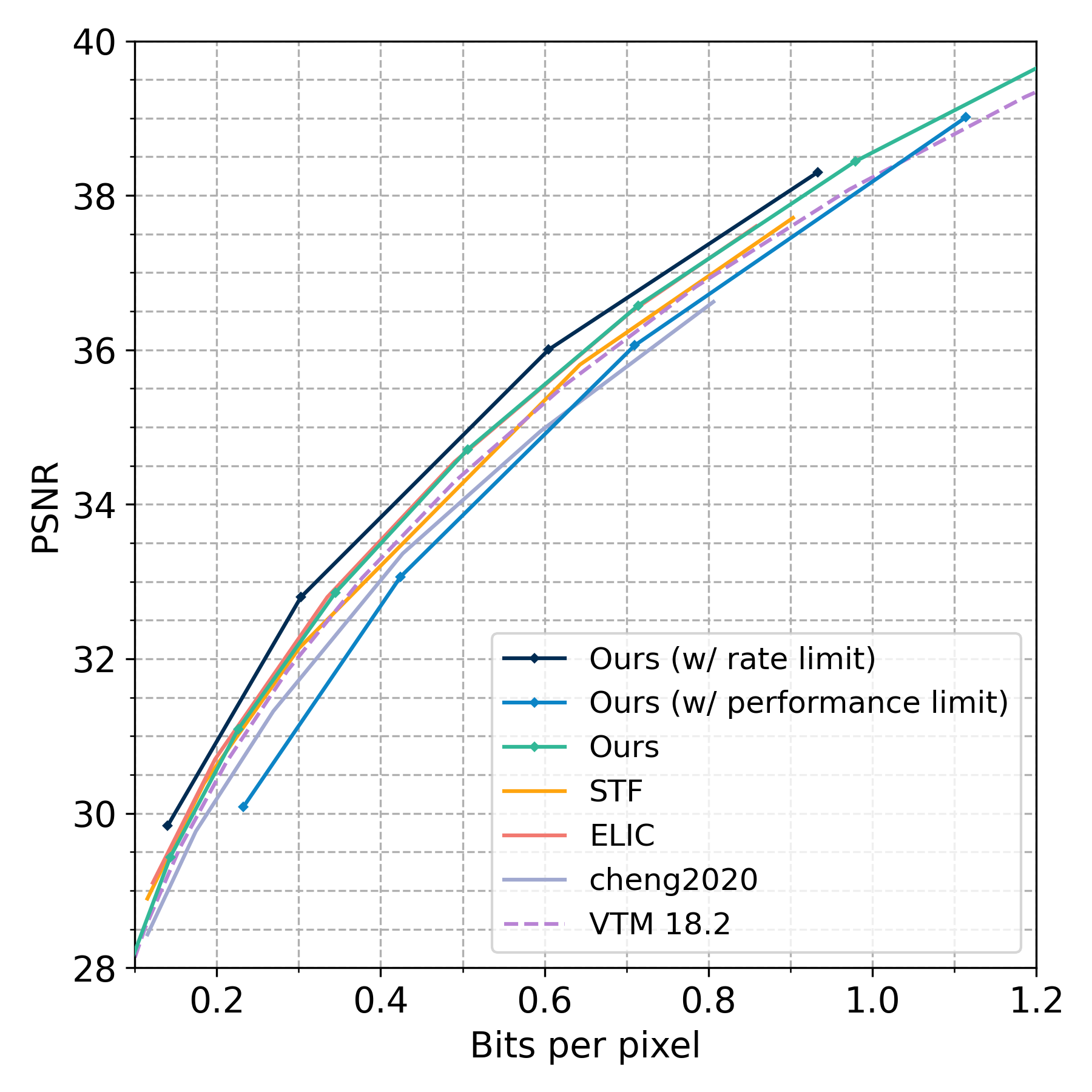}
        \vspace{-6mm}
        \subcaption{}
    \end{minipage}
    \begin{minipage}{0.35\linewidth}
        \centering
        \includegraphics[width=\linewidth]{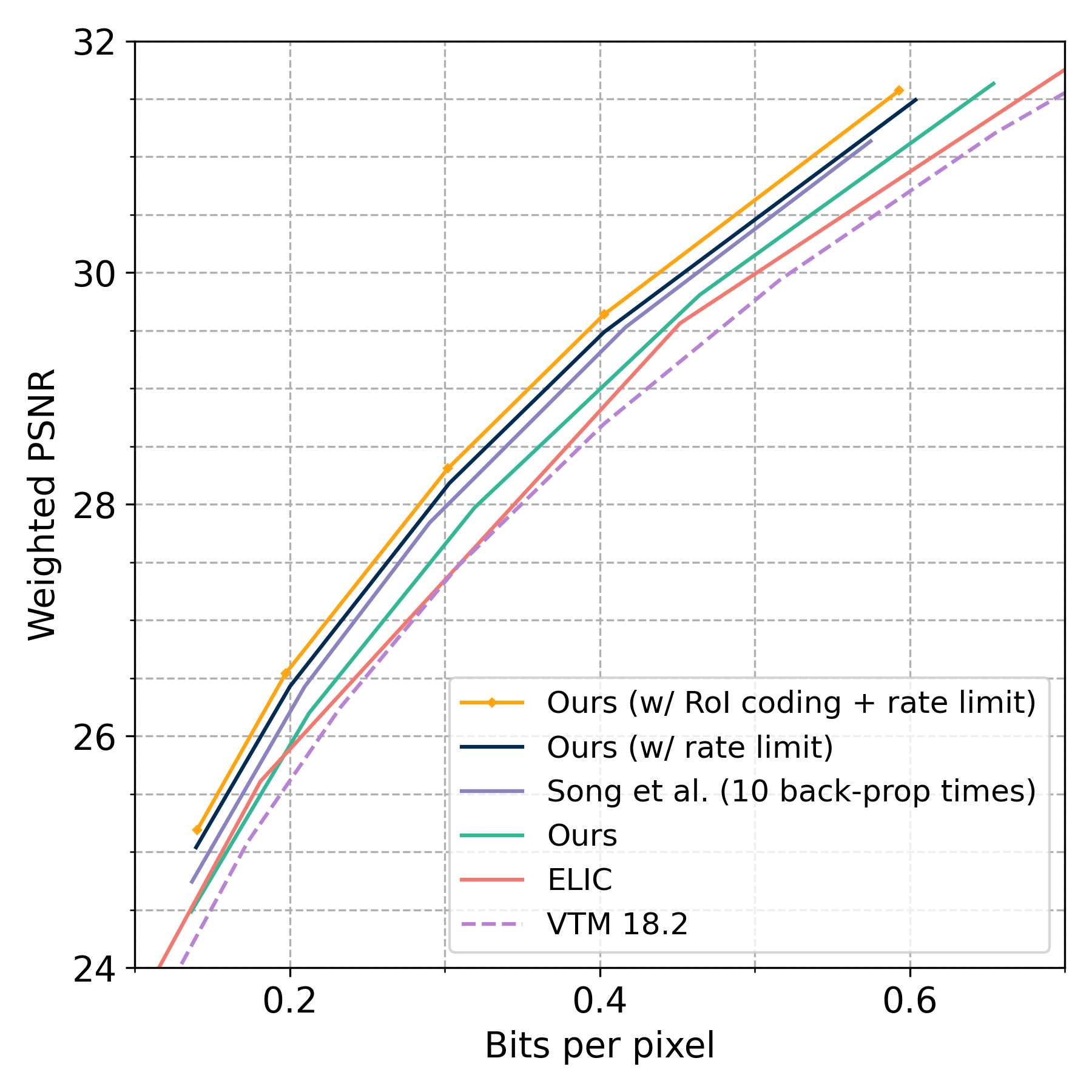}
        \vspace{-6mm}
        \subcaption{}
    \end{minipage}
    \vspace{-2mm}
    \caption{The effectiveness of refinement of our Comp-X.}
    \vspace{-2mm}
    \label{fig:refinement}
\end{figure}
\myparagraph{Effectiveness of refinement}\quad
We validate the refinement capability of our Comp-X under various constraints, including both simple and hard requests.
For simple request, we evaluate tasks that aim to minimize distortion while adhering to either a rate or performance constraint. As shown in Fig.~\ref{fig:refinement}(a), effective instance-level rate control in "Ours (w/ rate limit)" leads to a significant improvement in RD performance compared to the baseline "Ours." Conversely, when a performance limit (\eg, PSNR) is enforced, "Ours (w/ performance limit)" requires a higher average bitrate to achieve the same performance as the baseline. 

For hard request, we use the instruction shown in the last row of Table~\ref{tab:test_instruction}, assuming the user is an engineer with a certain level of expertise. In this scenario, Comp-X dynamically balances the quality factors between RoI and non-RoI regions, aiming to maximize weighted PSNR while satisfying the file size constraint. The results are illustrated in Fig.\ref{fig:refinement}(b). Notably, Song~\etal~\cite{song2021variable} updates a learnable spatial-wise quality map through gradient backpropagation using a specific rate-distortion loss. For a fair comparison, we also perform 10 optimization iterations on the map, using a weighted MSE loss function: $\lambda \times (0.8 \text{MSE}{\text{RoI}} + 0.2 \text{MSE}{\text{NRoI}}) + R$. Ours (w/ RoI coding + rate limit) demonstrates superior performance, showcasing the ability of our proposed Comp-X to adaptively adjust multiple parameters to satisfy complex constraints.


\section{Conclusion}
We introduced Comp-X, the first interactive learned image compression paradigm utilizing large language model (LLM) agents to automate image compression tasks based on user instructions. We conduct it through the following three innovations: (1) We build a unified image compression framework integrating diverse coding modes, including human-machine perception, variable coding, and spatial bit allocation. (2) We propose an augmented in-context learning method with coding expert feedback to teach the LLM agent how to understand the coding request, mode selection, and the use of the coding tool. (3) We propose the first benchmark designed for interactive image compression, featuring diverse user requests with expert annotations.
Experiments show that Comp-X effectively understands coding requests, delivers strong textual interaction capabilities, and maintains competitive compression performance, providing a promising avenue for artificial general intelligence (AGI) in image compression. 

{
    \small
    \bibliographystyle{ieeenat_fullname}
    \bibliography{main}
}


\clearpage
\appendix

\section{More Details of IIC Bench}
\subsection{Examples for simple task}
\begin{figure}[htbp]
\centering
\begin{verbatim}
"instruction": "Keep the screen in image.png clear.",
"label": {
    "file_path": "image.png",
    "compression_mode": "distortion",
    "RoI_coding": true,
    "RoI_object": "screen",
    "Object_needed_to_be_transmitted": "all",
    "encoded_size_level": "medium",
    "specific_performance_limit": false,
    "specific_bitrate_limit": false,
    "performance_metric": "weighted_PSNR(0.8, 0.2)",
    "bitrate_min": null,
    "bitrate_max": null,
    "bitrate_unit": null,
    "performance_min": null,
    "performance_max": null
}
\end{verbatim}
\begin{verbatim}
"instruction": "Make sure picture.jpg looks great after compression.",
"label": {
    "file_path": "picture.jpg",
    "compression_mode": "perception",
    "RoI_coding": false,
    "RoI_object": null,
    "Object_needed_to_be_transmitted": "all",
    "encoded_size_level": "medium",
    "specific_performance_limit": false,
    "specific_bitrate_limit": false,
    "performance_metric": "perception",
    "bitrate_min": null,
    "bitrate_max": null,
    "bitrate_unit": null,
    "performance_min": null,
    "performance_max": null
}
\end{verbatim}
\begin{verbatim}
"instruction": "Only compress the background in this picture.png.",
"label": {
    "file_path": "picture.png",
    "compression_mode": "distortion",
    "RoI_coding": false,
    "RoI_object": null,
    "Object_needed_to_be_transmitted": "background",
    "encoded_size_level": "medium",
    "specific_performance_limit": false,
    "specific_bitrate_limit": false,
    "performance_metric": "distortion",
    "bitrate_min": null,
    "bitrate_max": null,
    "bitrate_unit": null,
    "performance_min": null,
    "performance_max": null
}
\end{verbatim}
\end{figure}

\newpage

\subsection{Examples for hard task}
\begin{figure}[h]
\centering
\begin{verbatim}
"instruction": "Compress the image.jpg for object detection and 
ensure the file size is under 500KB."
"label": {
    "file_path": "image.jpg",
    "compression_mode": "detection",
    "RoI_coding": false,
    "RoI_object": null,
    "Object_needed_to_be_transmitted": "foreground",
    "encoded_size_level": "medium",
    "specific_performance_limit": false,
    "specific_bitrate_limit": true,
    "performance_metric": "detection",
    "bitrate_min": null,
    "bitrate_max": 500,
    "bitrate_unit": "KB",
    "performance_min": null,
    "performance_max": null
}
\end{verbatim}
\begin{verbatim}
"instruction": "Compress report.jpg to ensure the text remains clear 
                and keep the file size under 100KB.
"label": {
    "file_path": "report.jpg",
    "compression_mode": "distortion",
    "RoI_coding": true,
    "RoI_object": "text",
    "Object_needed_to_be_transmitted": "all",
    "encoded_size_level": "medium",
    "specific_performance_limit": false,
    "specific_bitrate_limit": true,
    "performance_metric": "weighted_PSNR(0.8, 0.2)",
    "bitrate_min": null,
    "bitrate_max": 100,
    "bitrate_unit": "KB",
    "performance_min": null,
    "performance_max": null
}
\end{verbatim}
\end{figure}

\begin{figure*}[t]
    \centering
    \includegraphics[width=\linewidth]{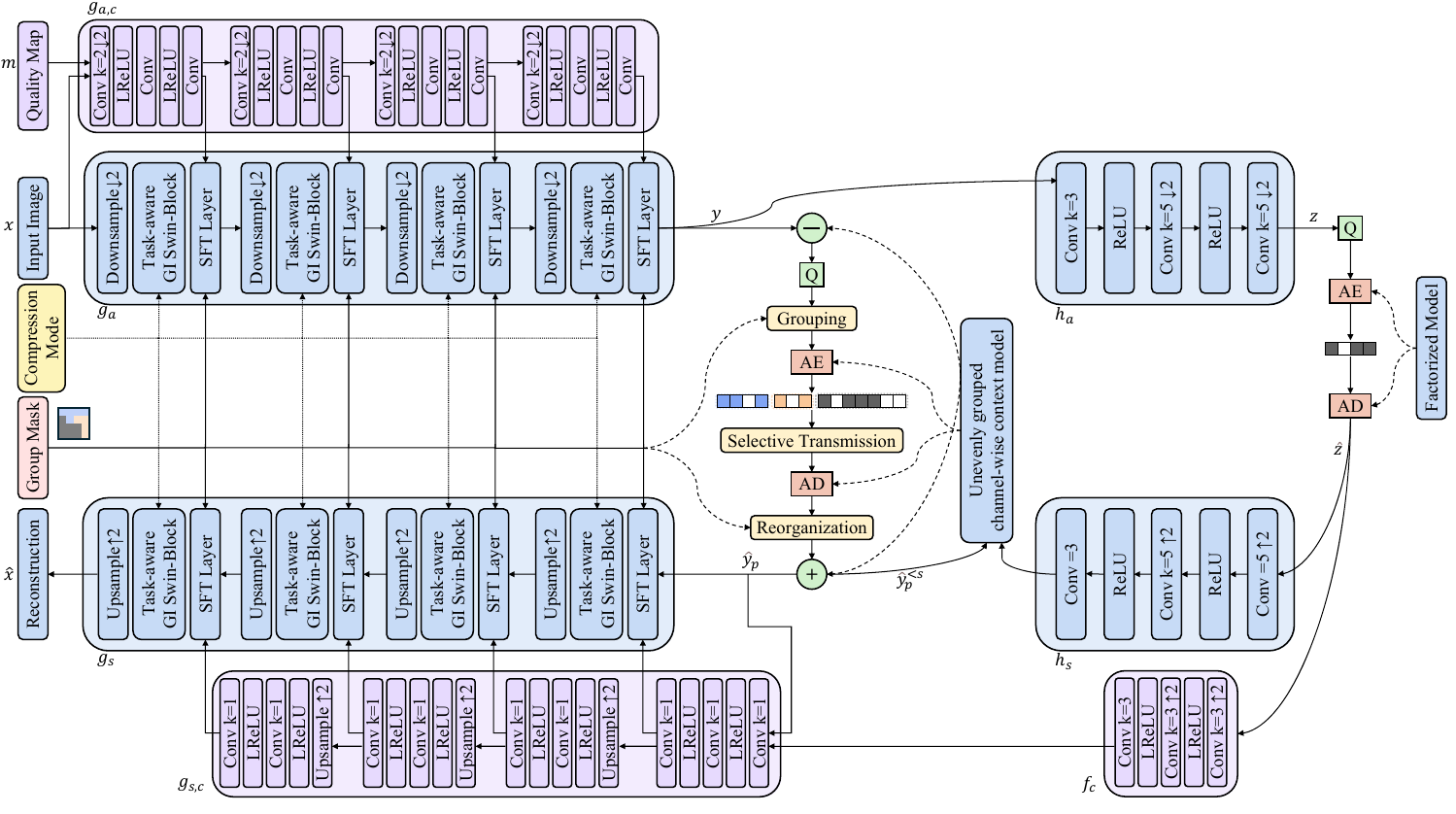}
    \caption{The network architecture of our multi-functional image codec.}
    \label{fig:appd_codec_details}
\end{figure*}

\section{More Details of Multi-functional Image Codec}

\begin{figure}[t]
    \centering
        \begin{minipage}{0.55\linewidth}
        \centering
        \includegraphics[width=0.8\linewidth]{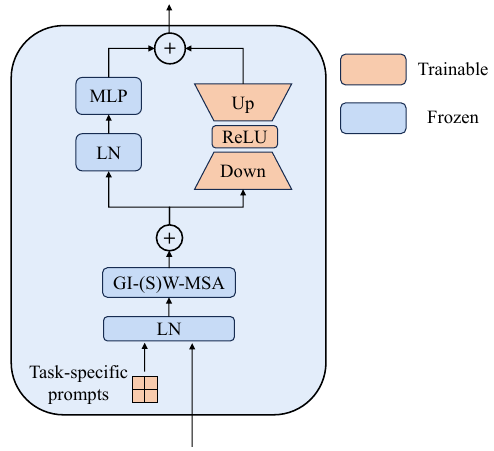}
        \caption{Task-aware GI Swin-Block.}
        \label{fig:appd_codec_GI_block}
    \end{minipage}
    \begin{minipage}{0.4\linewidth}
        \centering
        \includegraphics[width=\linewidth]{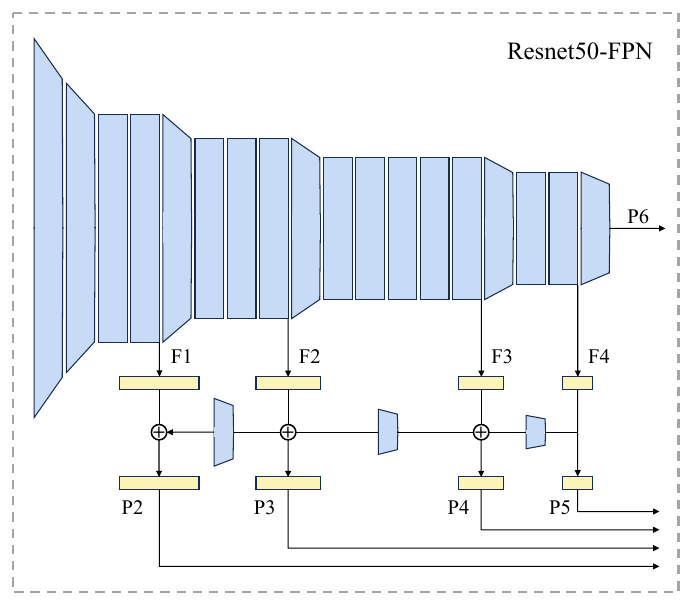}
        \caption{Architecture of ResNet50-based FPN.}
        \label{fig:appd_resnet50}
    \end{minipage}
    \hfill
\end{figure}

\paragraph{Network architecture}\quad
Fig.~\ref{fig:appd_codec_details} illustrates the architecture of our multi-functional image codec. The main transformation is performed in a staggered manner by integrating upsampling/downsampling operations, Task-aware Group-Independent Swin Blocks (Task-aware GI Swin-Blocks), and Spatial Feature Transform (SFT) layers. Specifically, the upsampling module consists of a pixel shuffle operation followed by a 1×1 convolution layer, while the downsampling module includes a pixel unshuffle operation and a 1×1 convolution layer. The GI Swin-Block, originally proposed by Feng~\etal~\cite{feng2023semantically}, maintains strong transformation capabilities while preserving independence among groups in the latent representations. To enable support for diverse tasks, we unify the codec using a multi-adapter approach that switches specific parameters based on the compression mode. Fig.\ref{fig:appd_codec_GI_block} shows our task-aware GI Swin-Block. We incorporate learnable task-specific prompts\cite{chen2023transtic} and a parallel learnable MLP branch~\cite{chen2022adaptformer} to adapt a pre-trained MSE-based codec for various machine vision tasks.

Following Song~\etal~\cite{song2021variable}, we integrate the SFT layer to support variable-rate compression and spatial bit allocation. The SFT layer uses the transformed feature of the quality map as a condition to generate scaling and shifting parameters, which are applied through an element-wise affine transformation on the input feature. This enables fine-grained control over the spatial, pixel-level quality of the image. 
To maintain group independence, we replace the original 3×3 convolution in the SFT layer~\cite{song2021variable} with 1×1 convolution layer. Additionally, the 3×3 convolution in the transform module of the quality map on the encoder side is replaced with a 2×2 convolution with stride step $=$ 2.

For entropy coding, we adopt the unevenly grouped channel-wise context model~\cite{he2022elic}. However, the convolutional transformations used by He~\etal~\cite{he2022elic} introduce spatial dependencies across groups, compromising the independence required for group-wise decoding. To address this, we replace their transformation module with the GI Swin-Block. We split $y$ along the channel dimension into five chunks with 16, 16, 32, 64, and 192 channels, respectively.

\myparagraph{Training details}\quad

We train our MSE-optimized variable-rate codec for 1.5M iterations, which is resumed from the hyperprior model in~\cite{feng2023semantically}. The loss function is:
\begin{equation}
\mathcal{L} = \sum_{i}^{N} \lambda_i \frac{(x_i - x'_i)^2}{N} + \mathcal{R}(\hat{y}) + \mathcal{R}(\hat{z})
\label{equ:app_rd}
\end{equation}
where $\mathcal{R}(\hat{y})$, $\mathcal{R}(\hat{z})$ represent the bitrates of $\hat{y}$ and $\hat{z}$ estimated by the entropy model, respectively. The weight $\lambda_i$ is determined by a corresponding quality map $m_i$ by a predefined and monotonically increasing function $T : [0,1] \rightarrow \mathbb{R}^+$, i.e., $\lambda_i = T(m_i)$, following~\cite{song2021variable}. Therefore, the higher quality level $m_i$ is, the higher weight $\lambda_i$ is for the corresponding pixel $x_i$ distortion term.

Next, we incorporate adapters to support various downstream tasks. 
For all the machine vision tasks, we set the channel dimension of learnable MLP branch to 128 and use 16 learnable task-specific tokens. The adapters are fine-tuned for 600K iterations using task-specific perceptual losses $d(x, \hat{x})$:
\begin{equation}
\mathcal{L} = \lambda d(x, \hat{x}) + \mathcal{R}(\hat{y}) + \mathcal{R}(\hat{z})
\end{equation}
Here, $\lambda$ shares the same meaning as $\lambda_i$ in Eq.~\ref{equ:app_rd}, but applied uniformly across all pixels, instead of being adapted per-pixel based on the quality map. Specifically, the perceptual loss is evaluated based on a pre-trained ResNet50~\cite{he2016deep}, Faster R-CNN~\cite{ren2015faster}, Mask R-CNN~\cite{he2017mask} and Keypoint R-CNN~\cite{he2017mask} for classification, object detection, instance segmentation and pose estimation, respectively. Fig.~\ref{fig:appd_resnet50} illustrates a ResNet50-based Feature Pyramid Network (FPN), which serves as the feature extractor in Faster R-CNN, Mask R-CNN and Keypoint R-CNN. For the classification task, the perceptual loss is evaluated in the feature space of F1, F2, F3, and F4:
\begin{equation}
d(x, \hat{x}) = \frac{1}{4} \cdot \sum_{l=1}^{4} MSE(F_l(x), F_l(\hat{x})),
\end{equation}
where $x$ and $\hat{x}$ are the input and decoded images, respectively. For the tasks of object detection, instance segmentation and pose estimation, the perceptual loss is evaluated in the feature space of P2, P3, P4, P5, and P6:
\begin{equation}
d(x, \hat{x}) = \frac{1}{5} \cdot \sum_{l=2}^{6} MSE(P_l(x), P_l(\hat{x})).
\end{equation}

For perception task, we fine-tune the adapter in the encoder and the main decoder ($g_s, g_{s,c}$). We set the channel dimension of learnable MLP branch to 256 and use 16 learnable task-specific tokens. 
We utilize the two-stage training process of HiFiC~\cite{mentzer2020high}. In the first stage, we train the model for 1M iterations without adversarial loss using the following loss function:
\begin{equation}
\mathcal{L}_{\text{1st}} = \lambda \left( \lambda_d \, d(x, \hat{x}) + \mathcal{L}_P(x, \hat{x}) \right) + \mathcal{R}(\hat{y}) + \mathcal{R}(\hat{z}),
\end{equation}
where $R$, $d$, $\mathcal{L}_P$ represent the bitrate, MSE, and LPIPS~\cite{zhang2018unreasonable}, respectively. We set $\lambda_d=150$.

In the second stage, we fine-tune the model with the adversarial loss~\cite{msillm} for 500k iterations as follows:
\begin{equation}
\mathcal{L}_{2nd} = \lambda ( \lambda_d d(x, \hat{x}) + \lambda_P \mathcal{L}_P(x, \hat{x}) + \lambda_{adv} \mathcal{L}^G_{\text{GAN}}) + \mathcal{R}(\hat{y}) + \mathcal{R}(\hat{z}),
\end{equation}
where $\lambda_{adv} \mathcal{L}^G_{\text{GAN}}$ represents the generator loss. We set $\lambda_d=150$ and $\lambda_{adv}=0.008$. 
To train the discriminator across different bitrates, we employ three independent discriminators that evenly span the entire bitrate range, following the method in CRDR~\cite{CRDR}.

We optimize the model using Adam~\cite{kingma2014adam}. We begin with linear warmup~\cite{Liu2020linearwarmup} for 10,000 steps and adopt a cosine learning rate decay for the rest of training. The peak learning rate is set to $4 \times 10^{-4}$ for training discriminators and $1 \times 10^{-4}$ for all other experiments. All experiments are conducted on an NVIDIA RTX 3090 GPU.

\myparagraph{Evaluation details}\quad
We evaluate the performances of pose estimation, instance segmentation, and object detection using Keypoint R-CNN~\cite{he2017mask}, Mask R-CNN~\cite{he2017mask} and Faster R-CNN~\cite{ren2015faster}, respectively.  All these models utilize the R50-FPN backbone~\cite{he2016deep} and are implemented using the Detectron2 toolbox~\cite{wu2019detectron2}. For classification evaluation, we use ResNet-50~\cite{he2016deep}. When jointly considering both human and machine vision, in addition to evaluating task performance, we also provide selective reconstruction quality for objects across all categories in instance segmentation tasks and for human-related objects in pose estimation tasks. For bits per pixel (bpp) is computed as \(\frac{b}{h \times w}\)
where \( h \) and \( w \) denote the original image resolution, and \( b \) represents the total bits in encoded bitstream.

\section{System Prompt for Planning}
\begin{tcolorbox}[
  breakable,
  colback=gray!5,
  colframe=gray!60!black,
  title=System Prompt,
  fonttitle=\bfseries,
  listing only,
  listing options={
    basicstyle=\ttfamily\footnotesize,
    breaklines=true,
    columns=fullflexible,
    upquote=true,
    escapeinside=||,
    keywordstyle=\color{blue},
    showstringspaces=false
  }
]
You are an intelligent assistant responsible for analyzing user input to extract parameters for an image compression task using a multi-functional image codec based on a semantic structured bitstream. Each part of the bitstream represents a specific object and can be directly used for various intelligent tasks. Your goal is to accurately parse the user's instructions and extract the necessary parameters for compression, ensuring the output aligns with the user's intent.
For scenarios that serve human vision like distortion and perception, we default to transmitting all objects in the image.
For scenarios that serve machine vision like classification, segmentation, or detection, we default to transmitting only task-relevant content ('foreground' or <specific object(s), if applicable>). In addition, you need to carefully determine whether the user indicates that only part of the image needs to be transmitted.
Sometimes, users need to balance both human and machine vision tasks. In such cases, you can set the compression mode to distortion while still transmitting only specific object(s) for machine vision tasks.

Based on the user's input, extract the following parameters:

1. Image file path (e.g., "example.png"). 

2. Compression mode:
- Options: distortion, perception, classification, segmentation, detection or pose estimation.
- Choose distortion if not explicitly specified.
- For downstream tasks not listed, default to perception.
- Mode Descriptions:
-- distortion: Minimizes distortion for general quality needs.
-- perception: Optimizes subjective visual quality for human perception.
-- classification: Preserves key features for image classification tasks.
-- segmentation: Retains details important for segmentation tasks.
-- detection: Prioritizes information for object detection tasks.
-- pose estimation: Focuses on features essential for pose estimation tasks.

3. Region of Interest (RoI)
- Check for a specific region of interest (RoI) that requires quality preservation, and specify the region if applicable.

4. Transmission Requirements
- Determine if only specific objects (e.g., "foreground" or "vehicles") need to be transmitted.
- Default Behavior:
-- For human vision tasks (e.g. distortion, perception), transmit all objects.
-- For machine vision tasks (e.g. classification, detection), transmit only task-relevant content.

5. RoI\_coding determines the evaluation metric or quality prioritization for a specific region but does not imply exclusion of other objects unless explicitly stated. If the user specifies "high quality" for certain objects but does not explicitly exclude other regions, transmit 'all' objects while prioritizing the quality of specified regions.

6. You are highly familiar with common units used to describe file sizes and metrics, such as Bytes (B), KB, MB, and PSNR (dB), among others.

7. Target file size recognition and bitrate limit:
- The default encoded\_size\_level is medium. You can adjust the encoded\_size\_level to meet the user's more ambiguous needs, such as "optimal distortion" or "as xx as possible." The larger the encoded\_size\_level, the higher the task performance.
- If the user specifies a target file size (e.g., "Target file size is 7000 Bytes"), set `specific\_bitrate\_limit=True` and set `bitrate\_max` accordingly.
- Ensure `bitrate\_unit` matches the unit specified by the user.
- If specific\_bitrate\_limit=true, prioritize the bitrate constraint over encoded\_size\_level. In such cases, set encoded\_size\_level to "medium".

8. Identify the performance metric:
- In most cases, this will correspond to the task or mode name (e.g., "distortion," "segmentation").
- If the user specifies a metric (e.g., "PSNR," "MS-SSIM"), use the specified metric.
- Special case:
-- If RoI\_coding=True and compression\_mode=distortion, evaluate using weighted\_PSNR.
-- If the user specifies a ratio for weighted\_PSNR, return the metric in the format weighted\_PSNR([ratio of the RoI region], [ratio of the non-RoI region]), ensuring the sum of both ratios equals 1, e.g. weighted\_PSNR(0.8, 0.2).

9. If the user has multiple requirements at the same time, set 'performance\_metric'=[metric1, metric2], such as "performance\_metric": ["perception", "weighted\_PSNR(0.8, 0.2)"].

Output format: Return the results in the following JSON structure. For optional parameters, use null or other appropriate default values where not applicable: \\
\{\\
  "file\_path": "<image file path>",\\
  "compression\_mode": "<compression mode>",\\
  "RoI\_coding": true/false,\\
  "RoI\_object": null or "foreground" or "<specific object class, if applicable>",\\
  "Object\_needed\_to\_be\_transmitted": "all" or "foreground" or "<specific object(s), if applicable>",\\
  "encoded\_size\_level": "medium" or "minimum" or "maximum",\\
  "specific\_performance\_limit": true/false,
  "specific\_bitrate\_limit": true/false,\\
  "performance\_metric": "<task/mode name>" or "<user-specified metric>",\\
  "bitrate\_min": null or <minimum bitrate value>,\\
  "bitrate\_max": null or <maximum bitrate value>,\\
  "bitrate\_unit": null or "<bitrate unit>",\\
  "performance\_min": null or <minimum performance value>,\\
  "performance\_max": null or <maximum performance value>\\
\}\\
Analyze the user's intent step by step to ensure the parameters are set correctly.
\end{tcolorbox}

\section{Case Study}
We provide a case study on the refinement process of RoI coding in Fig.~\ref{fig:refinement} to help readers better understand
the process of our Comp-X. To precisely meet the user constraint, if only an upper bitrate limit is specified, we set the lower limit to 256 bytes below it.

\begin{tcolorbox}[
  breakable,
  colback=orange!5!white,       
  colframe=orange!30!white,      
  title=Case Study on Refinement Process,
  fonttitle=\bfseries,
  coltitle=black,
  listing only,
  listing options={
    basicstyle=\ttfamily\footnotesize,
    breaklines=true,
    columns=fullflexible,
    upquote=true,
    escapeinside=||,
    keywordstyle=\color{blue},
    showstringspaces=false
  }
]

User: Compress kodim03.png. Keep foreground objects with high quality. When evaluating the result, I want to use weighted PSNR, and set the RoI region scale to 0.8. A Target file size is about 15000 Bytes.\\

Comp-X [Planning]: \\
\{\\
'file\_path': 'kodim03.png',\\
'compression\_mode': 'distortion',\\
'RoI\_coding': True,\\
'RoI\_object': 'foreground',\\
'Object\_needed\_to\_be\_transmitted': 'all',\\
'encoded\_size\_level': 'medium',\\
'specific\_performance\_limit': False,\\
'specific\_bitrate\_limit': True,\\
'performance\_metric': 'weighted\_PSNR(0.8, 0.2)',\\
'bitrate\_min': None,\\
'bitrate\_max': 15000,\\
'bitrate\_unit': 'Bytes',\\
'performance\_min': None,\\
'performance\_max': None\\
\}\\

Comp-X [Execution]: \\
iteration 0, q\_factor: [0.5, 0.5],\\
bytes: 5255, performance: 28.6174\\

Comp-X [Refinement]: \\
Thoughts: The current bitrate in bytes is 5255, which is significantly below the required range of 14744 to 15000 bytes. Therefore, the current q values do not meet the user's requirements for bitrate. We need to increase the q values to increase the file size while keeping the weighted\_PSNR performance metric in mind. Since the metric is weighted\_PSNR(0.8, 0.2), we should try to increase the first q value more than the second one.

q= [0.8, 0.6]\\

Comp-X [Execution]: \\
iteration 1, q\_factor: [0.8, 0.6],\\
bytes: 10254, performance: 31.9122\\

Comp-X [Refinement]: \\
Thoughts: The current q values do not produce a bitrate within the required range of 14744 to 15000 bytes. The first attempt with q=[0.5, 0.5] resulted in a much lower bitrate (5255 bytes), while the second attempt with q=[0.8, 0.6] produced a higher performance but still below the required bitrate range (10254 bytes). Given that weighted\_PSNR(0.8, 0.2) suggests prioritizing the first q value, increasing it might help achieve the desired bitrate. We need to increase both q values to approach the target bitrate and maintain distinct values for non-RoI and RoI regions.

Suggest new q values:
q=[0.9, 0.7]\\

Comp-X [Execution]: \\
iteration 2, q\_factor: [0.9, 0.7],\\
bytes: 14676, performance: 33.4867\\

Comp-X [Refinement]: \\
Thoughts: The current q values [0.9, 0.7] result in a bitrate of 14676 bytes, which is just below the lower limit of 14744 bytes. The performance metric is weighted\_PSNR(0.8, 0.2), which suggests giving more importance to the non-RoI region. The current performance is 33.48665714. While the performance is good, the bitrate needs to be adjusted to meet the constraints.

Since we need to increase the bitrate slightly while maintaining or improving performance, we should increase the q value for the non-RoI region more than the RoI region to match the weighted importance.

Suggesting a new q value:
q=[0.91, 0.75]\\

Comp-X [Execution]: \\
iteration 3, q\_factor: [0.91, 0.75],\\
bytes: 16567, performance: 34.0238\\

Comp-X [Refinement]: \\
Thoughts: The user requirements specify that the bitrate should be between 14744 and 15000. The performance metric is weighted\_PSNR(0.8, 0.2), which suggests that the quality in non-RoI regions (weighted at 0.8) is more important. The most recent q value [0.91, 0.75] results in a bitrate of 16567 bytes, which exceeds the upper limit of the required range. Therefore, the current q value does not meet the user's requirements. To reduce the bitrate, we should decrease both q values, focusing more on the first q value since it has a larger weight.

Suggest a new q value:
q=[0.89, 0.74]\\

Comp-X [Execution]: \\
iteration 4, q\_factor: [0.89, 0.74],\\
bytes: 14676, performance: 33.8101\\

Comp-X [Refinement]: \\
Thoughts: The user requirements specify that the bitrate should be between 14744 and 15000 bytes. The current q values for the 2nd iteration ([0.9, 0.7]) result in a bitrate of 14676 bytes, which is slightly below the required range. The subsequent q values ([0.91, 0.75] and [0.89, 0.74]) produce bitrates exceeding the maximum limit. Therefore, the q values need adjustment to fall within the specified range. Additionally, the performance metric is weighted\_PSNR(0.8, 0.2), suggesting a higher weight on non-RoI regions. To better meet the requirements, I'll suggest slightly increasing the q values from the 2nd iteration while ensuring they remain distinct.\\

Comp-X [Execution]: \\
iteration 5, q\_factor: [0.905, 0.705],\\
bytes: 14919, performance: 33.5476\\

Comp-X [Refinement]: \\
Thoughts: The current q values [0.905, 0.705] result in a byte size of 14919, which is within the specified range of 14744 to 15000 bytes. The performance metric, weighted\_PSNR at 33.54759932, does not have a specified limit and is relatively high compared to previous attempts. Given that the bitrate is within the user requirements and the performance is optimized, parameter tuning can be stopped.

\end{tcolorbox}

\end{document}